# Case Study of a highly automated Layout Analysis and OCR of an incunabulum: 'Der Heiligen Leben' (1488)

Christian Reul
University of Würzburg
Am Hubland
D-97074 Würzburg
christian.reul@uni-wuerzburg.de

Marco Dittrich
University Library of Würzburg
Am Hubland
D-97074 Würzburg
marco.dittrich@uni-wuerzburg.de

Martin Gruner
University Library of Würzburg
Am Hubland
D-97074 Würzburg
martin.gruner@uni-wuerzburg.de

## ABSTRACT
This paper provides the first thorough documentation of a high quality digitization process applied to an early printed book from the incunabulum period (1450-1500). The entire OCR related workflow including preprocessing, layout analysis and text recognition is illustrated in detail using the example of 'Der Heiligen Leben', printed in Nuremberg in 1488. For each step the required time expenditure was recorded. The character recognition yielded excellent results both on character (97.57%) and word (92.19%) level. Furthermore, a comparison of a highly automated (LAREX) and a manual (Aletheia) method for layout analysis was performed. By considerably automating the segmentation the required human effort was reduced significantly from over 100 hours to less than six hours, resulting in only a slight drop in OCR accuracy. Realistic estimates for the human effort necessary for full text extraction from incunabula can be derived from this study. The printed pages of the complete work together with the OCR result is available online[1] ready to be inspected and downloaded.

## Categories and Subject Descriptors
Applied computing: document management and text processing – *document analysis, optical character recognition*.

## 1. INTRODUCTION
The amount of freely available scans of a variety of books increased massively in recent years. However, in order to make the underlying information available to researchers the scans have to be converted into machine-actionable text via Optical Character Recognition (OCR). While OCR on modern texts is (for the most part) seen as a solved task, an effective way of processing early printed books (mostly before 1500) was considered impossible only several years ago. The main reason was the absence of standardized typefaces that causes standard off the shelf OCR software to fail. Moreover, the overall quality of the scans (dirt, bleed-through, bad contrast, etc.) as well as irregular layouts (bordures, ornaments, marginalia, etc.) represent hurdles that have to be overcome.

Mainly due to the introduction of Long Short-Term Memory (LSTM) networks for OCR by Breuel et al. (see [1]) it is now possible to train models in order to recognize even the earliest printed books with accuracies of well over 95% (see [2] and [3]). While this seems to have solved the main problem, the other obstacles mentioned above still remain. Especially, segmentation seems to be the bottleneck when building towards a high throughput workflow eligible for mass digitization of early printed books. Fully manual methods offer very precise results but require many hours of human effort. In contrast, (semi-) automatic approaches can drastically cut down the time expenditure. Nonetheless, a quickly achieved segmentation is useless if it is not exact enough to achieve a satisfying subsequent OCR result. Optimizing this trade-off is an important subject while working towards high volume digitization of early printed books.

In the context of the *KALLIMACHOS* project[2] the digitization center of the University Library of Würzburg has implemented an effective digitization workflow. The goal of this study is to document the developed approach using the example of the incunabulum 'Der Heiligen Leben' (I.t.f.954)[3]. Additionally, insight on the tools, techniques and resources necessary to efficiently extract a high quality OCR result from the scanned pages of an early printed book is provided. The time spent on every step of the digitization pipeline is recorded. Moreover, the effects of a heavily automatic segmentation approach are examined. Therefore, the book got segmented two times using completely different approaches: fully manually and highly automated. The results are compared regarding time expenditure and obtained OCR accuracy.

The remainder of the paper is structured as follows: After a short review of related work in chapter 2 the task description is provided in chapter 3. As the workflow is separated into segmentation and OCR, these topics are covered in chapter 4 and 5 respectively. In chapter 6 the proposed methods are evaluated. Chapter 7 discusses the obtained results and chapter 8 concludes the paper.

## 2. RELATED WORK
This chapter briefly introduces a selection of tools and approaches related to OCR workflows. After introducing three OCR-engines and discussing their segmentation capabilities several layout analysis tools are covered. Finally, two promising approaches dealing with the OCR of early printed books are presented.

On books printed later than ~1900 the fee-based and proprietary *ABBYY-OCR*-engine[4] defines the state-of-the-art for layout analysis and OCR. However, as earlier typefaces are not included in the training models, the application on early printed books usually does not yield satisfactory results. Furthermore, it is impossible for the end user to influence the results by adapting to a given book, e.g. by training a document-specific model.

---

[1] https://go.uniwue.de/itf954ocr

[2] http://www.kallimachos.de

[3] Incunabula typographica folio Nr. 954. Available at: https://go.uniwue.de/itf954

[4] http://finereader.abbyy.com

The open-source OCR tool *Tesseract*[5] uses individual glyphs rather than complete text lines for training and recognition. Yet, recently released version 4.0 alpha added a new OCR engine based on LSTM neural networks. Besides standard pre-processing methods, Tesseract offers a built-in page layout analysis. Though, only box segmentation is supported and on complex layouts the results are not satisfying.

*OCRopus*[6] is an open-source OCR engine which allows the user to comfortably train and apply new models. Its approach is based on bidirectional LSTM networks that require images of an entire text line as input. Moreover, OCRopus provides pre-processing functionality such as binarization, deskewing and line segmentation. However, the layout analysis and text/non-text detection capabilities currently cannot match the state of the art.

In addition to a wide variety of other functions, the page segmentation tool *Aletheia* (see [4]) offers a very extensive set of fully manual and assisted segmentation features, allowing an almost pixel-perfect region annotation. Nonetheless, the automatic layout analysis is performed by just putting the Tesseract page segmentation module to use. Therefore, the user usually has to start from scratch, making the segmentation a very time consuming task. Furthermore, most of Aletheia's functionality is not available in the free-of-charge Lite version. In this paper the term 'fully manually' referring to Aletheia means that the user has to actively control the segmentation.

*Agora* developed by Ramel et al. [5] is quite similar to the LAREX tool used during this study. It is an interactive document image analysis tool which utilizes connected components in order to set up two maps for foreground and background. The user can interactively specify so called scenarios consisting of rules to adapt to a given document.

*SCRIBO* (see [6]), a module of the OLENA platform, is an open-source layout analysis framework. While not being a stand-alone tool, its modularity and flexibility allow for an easy integration into existing projects and workflows.

Springmann et al. (see [2] and [3]) showed that it is possible to consistently achieve character recognition rates above 95% even for the earliest printed books. Additionally, they introduced the idea of so-called mixed models which are trained on a variety of books with different typefaces. Applying these models to books that have not been part of the training corpus led to accuracy rates well above 90%. Future digitization projects might benefit massively from this, as it allows to start off by correcting a promising OCR result rather than starting from scratch in order to train a document-specific model.

Another interesting approach was proposed by Kirchner et al. (see [7]) who capitalized from the fact that printing houses commonly produced lots of books by using the same typefaces. Ideally, the obtained typesets can be applied to plenty of books after being trained once by utilizing EMOP's Franken+[7], yielding character accuracies of up to 95%.

## 3. TASK DESCRIPTION

The incunabulum I.t.f.954 of the University Library of Würzburg, is one of the most popular spiritual books of the Middle Ages. It is called the 'Legenda sanctorum' of Jacobus de Voragine in German translation as 'Der Heiligen Leben' printed by Anton Koberger in Nuremberg in 1488. Chronologically arranged according to the course of the church year, the text offers a collection of life stories of the saints, folk sacred legends and tracts of the church festivals.

I.t.f.954 is one of several books produced in Koberger's printing house which is part of the inventory of the University Library of Würzburg. Most of these books have a similar layout and an equally challenging typeset as I.t.f.954. Therefore, it should be a suitable test case to digitize Koberger's editions by applying the same segmentation and OCR techniques.

Figure 1 shows a representative selection of layouts occurring on the 792 pages of the book.

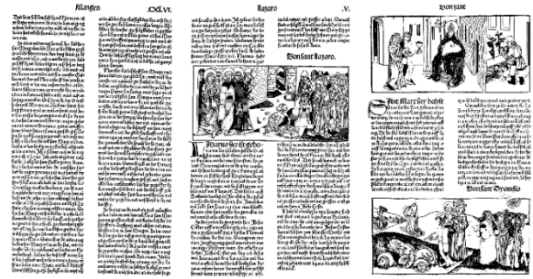

**Figure 1. Example pages from 'Der Heiligen Leben'.**

### 3.1 Requirements

For the purposes of the University Library of Würzburg it is often not sufficient to simply segment the scanned pages into text and non-text areas.

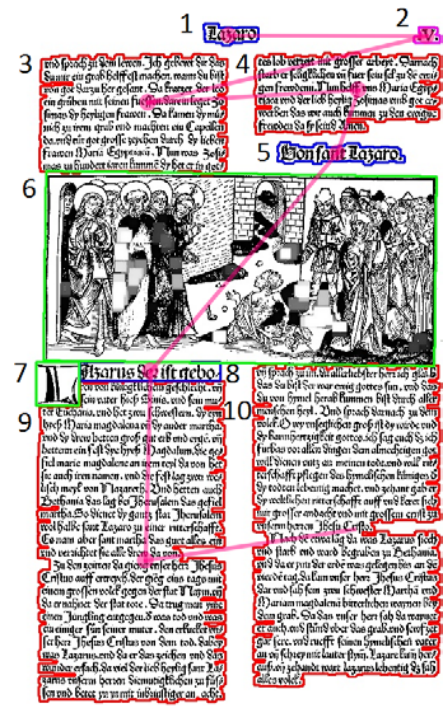

**Figure 2. Correct segmentation of an example page including images (6, 7) running text (3, 4, 9, 10), headings (1, 5, 8), the page number (2) and the reading order (polygonal chain).**

---

[5] https://github.com/tesseract-ocr/tesseract

[6] https://github.com/tmbdev/ocropy

[7] http://emop.tamu.edu/outcomes/Franken-Plus

For I.t.f.954 a detailed segmentation (see Figure 2) consisting of the following points was required:

- Accurate semantic distinction of region types: image, running text, heading and page number.
- A reading order that includes all text regions on a page.
- The so called *heading* segments can occur pretty much anywhere on the page and stretch over more than one line. In that case, they are supposed to be merged.
- Ornate initials have to be classified as images.

## 4. LAYOUT ANALYSIS

The segmentation process using Aletheia has proven to be a very precise but also quite cumbersome and tedious task. This chapter introduces an alternative open-source tool named LAREX and explains its application to I.t.f.954 in detail. Beforehand the preprocessing of the scanned pages is illustrated.

### 4.1 Preprocessing of the Scans

The initial 400dpi color scans were provided by the University Library of Würzburg. During a first preprocessing step the images were converted to binary using the Sauvola algorithm which also eliminated some stains. In a second step standard image processing techniques like dilation and erosion were used to remove the border between page and scan background, as well as the unwanted text block from the previous or next page respectively. The entire process was fully automated. Figure 3 shows an example result.

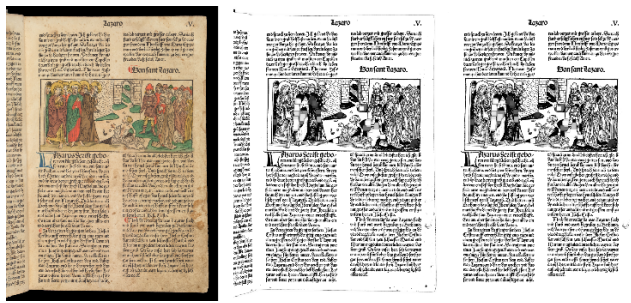

**Figure 3. Applied pre-processing steps: original scan (left), Sauvola output (middle), final result (right).**

### 4.2 LAREX

LAREX[8] (Layout Analysis and Region EXtraction) is a tool for interactive segmentation with type classification of scanned pages. It uses a rule-based connected components approach. A more in-detail description of the tool is provided in Reul et al.: LAREX – A semi-automatic open-source Tool for Layout Analysis and Region Extraction on Early Printed Books (submitted to this conference).

### 4.3 Application of LAREX to I.t.f.954

The segmentation workflow consisted of the following steps which will be discussed briefly in the upcoming sections:

1. Setup of the segmentation parameters.
2. Fully Automatic segmentation.
3. Manual correction of segmentation errors.

The standard functionality of LAREX had to be slightly extended in order to allow fully automatic segmentation.

#### 4.3.1 Setup of the Segmentation Parameters

The segmentation parameters were specified by processing a small number of pages semi-automatically. The resulting setup was very simple: images and paragraphs can occur anywhere, headings are expected to be located in the center at the top of the page and the page number can only appear in the top right corner. No parameters were changed during the remainder of the segmentation process.

#### 4.3.2 Fully Automatic Segmentation

The segmentation process is divided into a coarse and a fine segmentation step. Figure 4 shows the acquired interim results as well as the input and final output for an example page.

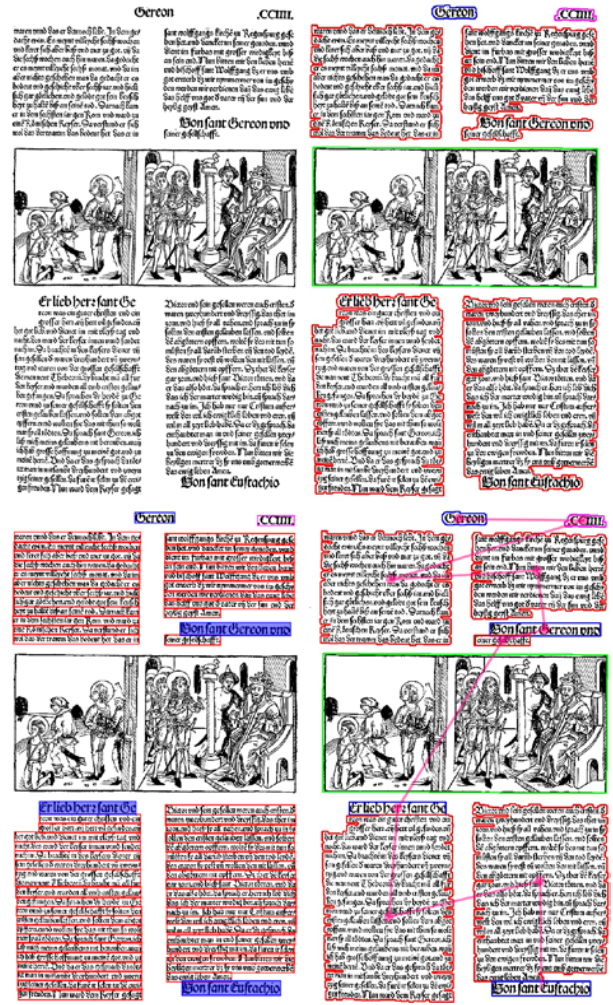

**Figure 4. Steps during the segmentation: Input (top left), coarse segmentation (top right), accepted heading candidates (blue, bottom left), final result (bottom right).**

During the fast coarse segmentation images are detected and connected regions are classified according to their respective area and position. However, in order to classify headings which do not occur at the top of the page or are still connected to another text block a more thorough segmentation is employed.

First, the text lines are detected by utilizing Tesseract's orientation and script detection and page segmentation functionality. To identify the headings a set of simple rules is applied: Only lines higher than the average height of all paragraph lines on the page are further tested. The Tesseract line segmentation merges two or even three lines into one on a regular basis. Therefore, the height of a line alone cannot be considered a sufficient criterion. However, the area of the connected components within a line, gives a good indication whether it is a heading or not. Thus, a candidate is marked as heading if the average area of its connected contours is at least 15% bigger than the overall average area of contours. As the size of the letters does not vary over the course of the book, this threshold was determined by using only a few pages.

---

[8] https://go.uniwue.de/larex

After detecting the *headings* the corresponding blocks are cut at the line borders if necessary and the types are assigned to the resulting segments. Two or more consecutive *heading* lines are merged. Lastly, the reading order is determined by applying simple rules. The results are stored using the PageXML standard.

*4.3.3 Manual Correction of Segmentation Errors*

Although the algorithm performs very well, all pages were manually checked and, if necessary, corrected. LAREX offers a variety of tools to perform manual amendments including the removal or splitting of regions and the changing of their types.

## 5. OCR USING OCROPUS

This chapter describes the necessary steps to process the obtained region annotation in order to extract machine readable text from the segments. The OCR workflow can be divided into the following four steps:

1. Extraction of the annotated regions.
2. Pre-processing and line segmentation.
3. Annotation and training.
4. Recognizing the text.

In addition to its text recognition capability, OCRopus offers several very useful built-in command line tools for pre-processing. The single steps are explained in detail throughout the following sections.

### 5.1 Extraction of the Annotated Regions

Independently of the segmentation method all regions have to be extracted and saved as separate images based on the information stored in the respective PageXML files. Earlier experiments have shown that the OCRopus text recognition performs slightly better on grayscale inputs than on binary ones. Therefore, the segments are extracted by using the original color scans. As the dimensions did not change during postprocessing, the segmentation obtained on the binary image is compatible with the original one (see Figure 5).

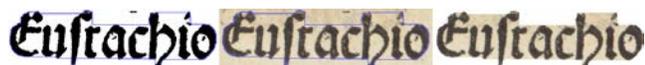

**Figure 5. Segmentation result achieved on binary (left), applied to original (middle) and extracted (right).**

### 5.2 Preprocessing and Line Segmentation

After the extraction the segments have to be deskewed and converted to binary in order to be prepared for the upcoming line segmentation. Part of the preprocessing is also the binarization via *ocropus-nlbin* which in addition to binary images generates normalized grayscale images for grayscale training and recognition. Then the line segmentation is performed on the individual image segments by applying *ocropus-gpageseg –gray* in order to extract both binarized and gray lines.

Since potential obstacles like swash capitals have already been removed during segmentation, the OCRopus line extraction usually performs very well.

### 5.3 Annotation and Training

In order to obtain some amount of ground truth for model training four pages were transcribed. This was considered sufficient, as in the case of I.t.f.954 four pages of plain text correspond to almost 400 lines containing over 2600 words. Apart from the segmentation (especially if performed manually) this is the most time consuming step of the OCR workflow. Different glyphs can look almost identical and their meaning may depend on the context. Furthermore, errors during the annotation phase can influence the OCR result heavily.

The training was performed by using about 75% of the annotated lines. To avoid overfitting training was stopped after 100 epochs (one epoch amounts to having statistically seen all 300 training lines). OCRopus stored a model after every 1,000 training steps. The best model was determined by testing on the remainder of annotated lines and selecting the one with the best recognition accuracy.

### 5.4 Recognizing the Text

With the best model identified the OCR on the entire book can now be conducted. All text lines get recognized one by one and are then concatenated according to the reading order defined during the segmentation step. Finally, to minimize pseudo-errors arising from white space around punctuation the OCR output was regularized by including blanks after punctuation characters in order to bring it in line with the transcription (ground truth) and minimize the required correction effort for the end user.

## 6. EVALUATION

In this chapter the measures of time expenditure of the entire workflow described above as well as the measured OCR accuracies are evaluated. Moreover, the results for the two different segmentation methods are compared.

### 6.1 Time Expenditure

In a practical application it is important to differentiate between tasks that are fully automated and tasks that require substantial human effort. In this study the segment extraction, pre-processing and line extraction, annotation and training, text recognition and the LAREX segmentation steps can be considered to be fully automated tasks whose runtime solely depend on computing power. Contrary, the setup of LAREX, the segmentation using Aletheia, the annotation of training data and the correction of the LAREX output were carried out manually. The required time-expenditure of the normalization is negligible.

The automatic tasks were performed on different PCs with similar hardware equipment (quad core processors and solid state disks). Student research assistants with extensive experience on the digitization of early printed books were responsible for the manual segmentation with Aletheia and the annotation of the training files. The setup of LAREX as well as the manual correction performed with it was conducted by the first author.

During the following sections three outcomes will be evaluated: manual segmentation using Aletheia, fully automatic segmentation using LAREX and the manually corrected LAREX output which was done within two hours.

While Table 1 displays the time exposures recorded for each of the fully automated steps described above, Table 2 focuses on the steps requiring human effort. All values have been rounded up to full hours.

Altogether (automated and manual), the workload of the complete workflow amounted to 125 hours (Aletheia) and 30 hours (LAREX, 32 hours including manual correction). The overall time expenditure was cut down to around 25% by utilizing LAREX during the segmentation. However, when only taking the human effort into account the time saving factor rises to 20 and even 34 depending on the LAREX output being manually corrected or not.

**Table 1. Recorded times for the individual steps that are performed fully automatically.**

| Task | Required time in h | |
|---|---|---|
| | Aletheia | LAREX |
| Segmentation including reading order | - | 4 |
| Region extraction | 4 | |
| Pre-processing and line segmentation | 4 | |
| OCRopus training | 10 | |
| Text recognition | 5 | |
| **Sum** | **23** | **27** |

**Table 2. Recorded times for the individual steps that require human effort.**

| Task | Required time in h | |
|---|---|---|
| | Aletheia | LAREX |
| Set-Up | - | 1 |
| Segmentation including reading order | 100 | - (+2) |
| Annotation for training | 2 | |
| **Sum** | **102** | **3 (+2)** |

As expected, the segmentation using Aletheia is the dominant factor. It has to be said that this value depends heavily on the individual demands of the user. During this study the guideline was to support the subsequent OCR as much as possible by excluding stains and other disturbing artefacts with justifiable expenditure.

## 6.2 OCR Accuracy

The accuracy of the character recognition was evaluated on 41 pages excluding the ones used for training and model selection by comparing the manually created ground truth to the OCR output. Table 3 shows the character accuracies for the four different inputs. In order to obtain nonparametric statistics the ISRI-Tools *accci* and *wordaccci* (see [8]) were used.

**Table 3. Character accuracy rates including the lower and upper limits of the approximate 95% confidence interval.**

| | Lower Limit | Mean | Upper Limit |
|---|---|---|---|
| Aletheia | 97.37% | 97.57% | 97.78% |
| Corrected | 97.10% | 97.37% | 97.64% |
| Full Auto | 97.08% | 97.35% | 97.63% |

The recognition of the manually segmented regions using Aletheia yielded the highest OCR accuracy with 97.57%. The fully automatically extracted regions by LAREX led to an accuracy of 97.35%. Interestingly, the manual correction of the LAREX segmentation improved the result only insignificantly to 97.37%. The calculated 95% confidence intervals of all methods except Aletheia include all other measured values.

The word accuracies shown in Table 4 lead to similar observations. All measured values are located within the 95% confidence intervals of the others and can therefore not be distinguished at this confidence level.

**Table 4. Word accuracy rates including the lower and upper limits of the approximate 95% confidence interval.**

| | Lower Limit | Mean | Upper Limit |
|---|---|---|---|
| Aletheia | 91.69% | 92.19% | 92.71% |
| Corrected | 91.16% | 91.75% | 92.35% |
| Full Auto | 91.22% | 91.84% | 92.47% |

## 6.3 Error Analysis

Despite the almost negligible difference between the OCR accuracies shown above this section tries to find possible explanations for the slight drop. First, some frequently occurring errors during the LAREX segmentation are examined. Then, the effect of unwanted inclusions is discussed.

### 6.3.1 LAREX Segmentation Errors

During the manual correction of the LAREX segmentation close to 80% of the pages did not require any manual amendment. The detection of heading lines, mostly caused by a flawed Tesseract line segmentation, represented the main source of error.

Another common problem are swash capitals as they are often very close or even directly connected to the rest of the adjacent text (see Figure 6). Therefore, the removal of the bounding rectangle of the detected contour can lead to a loss of characters. Furthermore, swash capitals that get overlooked almost certainly cause the subsequent line segmentation to fail.

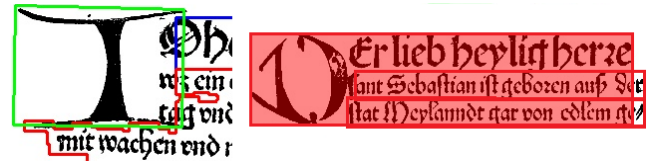

**Figure 6. Errors during swash capital removal: Lost characters because of initial removal (left); Missed initial causing the line segmentation to fail (right).**

Unneeded regions mostly caused by dirt or scan artefacts also occurred quite frequently but were very easy to remove and pretty much never influenced the rest of the segmentation. The assignment of the reading order as well as the detection of the images worked very reliably and caused almost no errors.

### 6.3.2 Effect of Unwanted Inclusions

One possible explanation for the drop in accuracy may be that the Aletheia segments are fitted tightly to the corresponding letters and words. On the contrary, the region borders created by LAREX leave some space due to the dilation operation applied before region detection. This can lead to additional speckles within the resulting segment (see Figure 7) which may disturb the OCR.

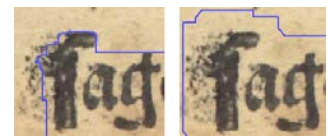

**Figure 7. Advantage of tightly fitted polygons regarding the insertion of unwanted speckles: Aletheia (left), LAREX (right).**

As LAREX cuts detected headings according to the recognized line around them, it is likely that the straight borders of the line sever or include parts of letters located within the adjacent lines (see Figure 8). Consequently, these incomplete letters or surplus parts of letters can lead to OCR errors (e.g. o → ō, m → in). When using Aletheia it is cumbersome but possible to include or exclude these parts.

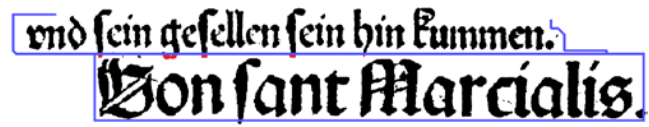

**Figure 8. Line segmentation causing the unwanted inclusion of parts of letters (red).**

## 7. DISCUSSION

The case study on the digitization of the incunabulum 'Der Heiligen Leben' confirmed recent research results claiming that OCR on even the earliest printed books is not only possible, but also very precise. 400 lines of transcribed ground truth sufficed in order to obtain a high-performance OCR model resulting in a character accuracy of 97.57% and a word accuracy of 92.19%.

Additionally, the experiment fortified the claim that layout analysis is the main bottleneck in an OCR workflow optimized towards larger volumes. The manual segmentation using Aletheia was responsible for 80% of the overall time expenditure. This number even rose to almost 98% when only considering the required human effort. The deployment of LAREX reduced the required time for segmentation significantly by achieving time saving factors exceeding 4 (overall) and 30 (manual effort).

Furthermore, it was shown that a slightly less precise segmentation only resulted in a minimal drop in OCR accuracy. A similar behavior is likely to occur when digitizing other early printed books. The manual correction of the LAREX segmentation results only led to a negligible improvement of the OCR accuracy. This implies that the coarse segmentation, especially text/non-text distinction, worked very reliably and most corrections became necessary because of semantic requirements. Therefore, an argument can be made that a fully automatic method leading to a quick, but not perfect segmentation result might often be the advisable approach. Consequently, the correction of semantic mistakes would have to be done on text tier, e.g. by using simple markups, instead of altering the segmentation results.

## 8. CONCLUSION AND FUTURE WORK

During this study it was shown that it is possible to obtain a high quality OCR result on an early printed book within a reasonable amount of time. Despite 'Der Heiligen Leben' comprising of almost 800 pages the entire OCR workflow, starting from the already scanned pages and resulting in the final OCR output, was completed in around 30 hours by investing less than five hours of human effort. The obtained character (over 97%) and word (around 92%) accuracies clearly illustrate OCRopus' ability to recognize even the earliest printed typefaces after being thoroughly trained.

Moreover, the benefits of a highly automated layout analysis tool like LAREX become very clear. Segmentation and post-correction remain the two dominant sources for required human effort during the digitization of early printed books. Thus, it is important to further optimize the inevitable trade-off between a precise segmentation and an accurate OCR result.

Our next project will be Vinzent of Beauvais'[9] 'Specula'. These 'encyclopaedias' of the Late Middle Ages seem to be tailor-made for another test run of the proposed workflow. Consisting of four books[10] the complete works comprise almost 3,300 pages (all available as high quality scans at the University Library of Würzburg) that for the most part possess a relatively homogeneous two column layout.

Hopefully, the capability of the proposed workflow can be documented even more clearly during this comprehensive and challenging task.

---

[9] Vincentius Bellovacensis, ca. 1190-1264.

## 9. ACKNOWLEDGMENTS
The authors would like to thank Uwe Springmann for providing insight and expertise throughout the paper, Phillip Beckenbauer, Sefika Karakaya and Maximilian Nöth for their work on I.t.f.954 as well as Hans-Günter Schmidt and Frank Puppe for their helpful remarks.

---

[10] Speculum doctrinale (1486), Speculum historiale (1483), Speculum morale (1485) and Speculum naturale (1481), all printed by Anton Koberger in Nuremberg.